\pdfoutput=1
\documentclass[11pt]{article}

\usepackage[]{acl}

\usepackage{times}
\usepackage{latexsym}

\usepackage[T1]{fontenc}
\usepackage[utf8]{inputenc}

\usepackage{microtype}

\usepackage{times}
\usepackage{latexsym}

\usepackage{times}
\usepackage{latexsym}
\usepackage{url}
\usepackage{amsfonts}
\usepackage{multirow}
\usepackage{amssymb}
\usepackage{subfloat}
\usepackage{graphicx}
\usepackage{arydshln}
\usepackage{amsmath}
\usepackage{caption}
\usepackage{CJKutf8}
\usepackage{caption}
\usepackage{capt-of}
\usepackage{amsmath}
\usepackage{CJKutf8}
\usepackage{amssymb}
\usepackage{pifont}
\usepackage{booktabs}
\usepackage{dsfont}
\usepackage{stmaryrd}

\title{Lost in Context? On the Sense-wise Variance \\
	of Contextualized Word Embeddings}

\author{Yile Wang \textnormal{and} Yue Zhang \\
	School of Engineering, Westlake University \\
	\texttt{\{wangyile,zhangyue\}@westlake.edu.cn} \\
}

\begin{document}
\maketitle
	
\begin{abstract}
	Contextualized word embeddings in language models have given much advance to NLP. Intuitively, sentential information is integrated into the representation of words, which can help model polysemy. However, context sensitivity also leads to the variance of representations, which may break the semantic consistency for synonyms. We quantify how much the contextualized embeddings of each word sense vary across contexts in typical pre-trained models. Results show that contextualized embeddings can be highly consistent across contexts. In addition, part-of-speech, number of word senses, and sentence length 	
	have an influence on the variance of sense representations. Interestingly, we find that word representations are position-biased, where the first words in different contexts tend to be more similar. We analyze such a phenomenon and also propose a simple way to alleviate such bias in distance-based word sense disambiguation settings.
	
\end{abstract}
	\section{Introduction}
Contextualized word embeddings (CWE) from pre-trained language models ~\cite{elmo,gpt2,bert,roberta,xlnet,albert,deberta} have achieved state-of-the-art performance in various downstream tasks. Different from static word embeddings, CWE represents words using a deep neural network, which can effectively model polysemy~\cite{elmo}, capture syntactic~\cite{hewitt-manning-2019-structural,wu-etal-2020-perturbed}, semantic~\cite{jawahar,Tenney}, and commonsense knowledge~\cite{factual,10.1162/tacl_a_00342} from large corpora.

However, research shows that context sensitivity can also bring unnecessary variation for CWE. For example, ~\citet{shi-etal-2019-retrofitting} show that the ELMo~\cite{elmo} embeddings of a word change drastically when the context is paraphrased, which makes the downstream model unrobust to paraphrasing and other linguistic variations. Such a phenomenon is beyond what we can expect from the perspective of semantic polysemy, which is not only undesirable linguistically but also has implications on NLP tasks such as word alignment~\cite{wordalign1,wordalign2} and lexical semantics~\cite{liu-etal-2020-towards,wang}. In the machine learning aspect, it has been shown that stable representations are relevant to reducing over-fitting, 
improving generalization and relieving data hunger~\cite{mccoy-etal-2020-berts,arora-etal-2020-contextual}.

Despite discussion on the consequence, there has been relatively little research characterizing the context-sensitivity in CWEs systematically. The only exception is Ethayarajh~\cite{ethayarajh-2019-contextual}, who provides overall statistics for ELMo, BERT, and GPT on the {\it token}-level. In this paper, we aim to quantify the variance in the \textit{sense}-level representations across contexts for different CWE models. Our goal is to quantitatively answer the following research questions. First, \textit{how much does CWE differs for the same word sense, and is the variation highly model-dependent?} Second, \textit{what types of words and contexts are more prone to CWE variations?} Third, \textit{how does word position affect CWE variations?} Knowledge of the above issues can be useful for better understanding and improving CWE, and also for guiding the choice of models toward a specific problem.

Empirically comparing seven pre-trained models (\textit{i.e.}, ELMo, BERT, SenseBERT, RoBERTa, DeBERTa, XLNet, and GPT2), we find:

(1) Word sense representation in CWE is generally consistent across contexts, while different models have varying degrees of sense-wise consistency.

(2) Sense-level representations vary differently according to word types and contexts, where they are more consistent for nouns or in short sentences.

(3) Sense-wise representation is position-sensitive for the sentence beginning words, where the first words in two sentences tend to be very close. We call this \textit{position bias} in CWE.

(4) We find that by add some simple prompt can largely change the distribution of CWEs, especially for the first words. These findings can be used for calibrating representation and are shown effective for a distance-based word sense disambiguation scenario.

To our knowledge, we are the first to investigate sense-level representation variances of CWE, and the first to report the influence of both token, context and position factors.

\section{Related Work}
Our work is in line with existing work on BERTology~\cite{rogers-etal-2020-primer}, which discusses the characteristics of pre-trained models, aiming to provide evidence for explaining their effectiveness as well as their limitations.

\textbf{Knowledge in CWE.} A line of research finds that CWE can encode transferable and task-agnostic knowledge and successfully improve the performance of downstream tasks~\cite{Nelson}. ~\citet{elmo} propose ELMo and find it encodes syntactic and semantic features at different layers. Goldberg~\cite{Goldberg2019AssessingBS} find BERT performs remarkably well on English syntactic phenomena. ~\citet{jawahar} reveal BERT composes a rich hierarchy of linguistic information, starting with surface features at the bottom, syntactic features in the middle followed by semantic features at the top. Our work is related to discussing knowledge of CWE, with a focus on word sense and context.

\textbf{Word Sense in CWE.} Recent work shows that CWE is capable of encoding word sense information. ~\citet{elmo-sense} find the ELMo  embeddings can be separated into multiple distinct groups, each with a certain meaning. ~\citet{vis} show different senses of a word are typically spatially separated using BERT representation. Garí Soler and Apidianaki~\cite{mono-poly} demonstrate that BERT representations reflect word polysemy level and their partition ability into senses. Such capability makes the success of leveraging CWE into sub-tasks such as word sense disambiguation~\cite{glossbert,yap-etal-2020-adapting,song-etal-2021-improved-word}. We also study sense-level knowledge inside CWE but differ in that our experiments are more on the \textit{variance} of embeddings for the \textit{same sense}, analyzing the influence from different contexts.

\textbf{Contextualization of CWE.} 
There have been works analyzing the variations of CWE according to different contexts. Ethayarajh~\cite{ethayarajh-2019-contextual} first compared ELMo, GPT2 and BERT, analyzing how the word representations change according to different models and layers. They find that the upper layers produce more context-specific representations, where the self-similarity of two identical words in different contexts becomes much lower. Our work is similar in the evaluation of  cross-sentence scenarios. However, we consider {\it sense}-level comparison rather than {\it word}-level comparison. \textit{Different} words in the \textit{same} sense are compared. In addition, we report not only the overall statistics for different models, but also discuss breakdown factors to influence context sensitivity, both internally and externally for each sense type.

~\citet{shi-etal-2019-retrofitting} find that ELMo cannot capture  semantic equivalence, where the word representations are very different in paraphrased contexts. To minimize the variance of CWE, they use paraphrases to retrofit word representation.
Similarly, ~\citet{elazar-etal-2021-measuring} claim that the invariance under meaning-preserving alternations in its input is a desirable property for CWE, however, they find the consistency is poor for some  factual knowledge, leading to the unstable knowledge representation. The above work does not consider sense-level representation and the comparison is restricted in the paraphrased context. In contrast, we use natural sentences with open context formation. In addition, we consider a wider range of models.

\section{Methods}
\label{section:method}

\subsection{Dataset and Models}

We use the XL-WSD dataset~\cite{pasini-etal-xl-wsd-2021}, an extra-large Word Sense Disambiguation benchmark, to investigate the CWE of words in the corpus with labeled sense information. Specifically, the \textsc{SemCor} and \textsc{WNGT} corpora are used, where the most widely used WSD dataset are gathered, including Senseval 2\&3~\cite{senseval2,senseval3} and SemEval 07\&13\&15~\cite{semeval07,semeval13,semeval15}. All sense taggings have been merged from various resources using BabelNet\footnote{\url{https://babelnet.org}}, which is a superset of Princeton WordNet, the standard sense inventory for English. Statistics of the dataset are shown in Table~\ref{table:statics}. 

We choose seven representative pre-trained models:

\textbf{ELMo}~\cite{elmo} uses a two-layer bi-directional LSTM~\cite{lstm} for CWE.

\textbf{BERT}~\cite{bert} makes use of bi-directional Transformer~\cite{transformer} and pre-trained with the mask language modeling and next sentence prediction. 

\textbf{SenseBERT}~\cite{sensebert} is based on BERT, but is pre-trained to predict both masked words and their WordNet supersenses.  

\textbf{RoBERTa}~\cite{roberta} is a variation of BERT, which applies more training corpora, dynamic masking, and removes the NSP objectives.

\textbf{DeBERTa}~\cite{deberta} is similar to RoBERTa, while using relative position embedding, disentangled attention mechanism and enhanced mask decoder. 

\textbf{XLNet}~\cite{xlnet} extends Transformer-XL~\cite{transxl} and is pre-trained using autoregressive permutation language modeling. 

\textbf{GPT2}~\cite{gpt2} uses a Transformer decoder structure for generative language model pre-training.

Specifically, we take the model checkpoints \textsc{ELMo-small}, \textsc{BERT-base-cased}, 
\textsc{RoBERTa-base}, \textsc{DeBERTa-base}, \textsc{XLNet-base-cased} and \textsc{GPT2-base} from AllenNLP~\cite{allennlp} and Huggingface~\cite{wolf-etal-2020-transformers}. For SenseBERT, we use the  \textsc{sensebert-base-uncased} released by ~\citet{sensebert}.  Most of the model outputs are in the same 768-dimensional vector space (except for 1024 dimensions for ELMo) so they can be fairly compared.

\begin{table}[t]
	\centering
	\small
	\begin{tabular}{l|c}
		\hline 
		\textbf{Types} &\textbf{Values} \\
		\hline 
		\#Total tokens &1,240,828 \\
		\#Labeled tokens  &273,815 \\	
		\#Vocabulary  &22,979 \\
		\hline
		\#Sentences  &83,277 \\
		\#Sentence length in average  &14.9 	\\
		\hline
		\#Senses inventory  &44,046 \\
		\#Senses for each token in average  &1.65  \\
		\#Maximum senses for one token  &58  \\
		\hline
	\end{tabular}
	\caption{Statistics of XL-WSD dataset of ~\citet{pasini-etal-xl-wsd-2021}.}
	\label{table:statics}
\end{table}

\subsection{Cosine Similarity}
For each word $t_i$ labeled with a specific sense $s_i$ and the corresponding context  $t_1$, ..., $t_i$, ..., $t_{N_i}$, we calculate the CWE $h^l_i$ using the above models:
\begin{equation}
h^l_{1}, ..., h^l_{i}, ..., h^l_{{N_i}} = \mbox{CWE} \ (t_1,..., t_i,..., t_{N_i})
\end{equation}
where $l$ is the model layer and $N_i$ is the sentence length. Special tokens such as [CLS] and [SEP] may be added according to different models while we discard them finally. We use its first sub-token vector for representation for those words that are split into multiple tokens~\cite{bpe,wordpiece}.

Following Ethayarajh~\cite{ethayarajh-2019-contextual}, we consider the normalized cosine similarity ${\rm Cos}(t_i,t_j) = \frac{<h_{{i}},h_{{j}}>}{\vert\vert h_{{i}}\vert\vert\cdot\vert\vert h_{{j}}\vert\vert}$ between two  words ${t_{i}}$ and ${t_{j}}$ with hidden CWE $h_i$ and $h_j$, respectively. Note that ${t_{i}}$ and ${t_{j}}$ may come from different sentences. For aggregating evaluation of embedding similarities in a corpus, we find all pairs of hidden representations from $t_i, t_j$, computing:
\begin{equation}
\texttt{Sim} = \frac{\sum_{i,j}{\rm Cos}(t_i,t_j)}{{\rm Count}(t_i,t_j)}
\label{equation:s1}
\end{equation}

We consider sense-wise variance for the following targets:

\textbf{The Same Word in the Same Sense.} We compare the representation of the same words in the same word sense. For each word pair, the representations should be similar intuitively. For example, the representation of word ``\textit{levels}''  should be similar when they share the same word sense \textsc{bn:00041239n} ({A relative position or degree of value in a graded group}) in the following sentences: 

$\bullet$ \textit{There are three \underline{\textbf{levels}} on which to treat the subject .}

$\bullet$ \textit{The dosages in the three \underline{\textbf{levels}} of the vertical profile were : ...}

\textbf{Different Words in the Same Sense.} Note that different words may have similar senses, showing that they share some common features in the semantic spaces. We also compare the CWE of different words but in the same word senses in certain contexts. For example, the words ``\textit{levels}'', ``\textit{layers}'' and ``\textit{strata}'' have the same word sense of \textsc{bn:00050303n} ({An abstract place usually conceived as having depth}) in the sentences below: 

$\bullet$  \textit{a good actor communicates on several \underline{\textbf{levels}}.}

$\bullet$  \textit{a simile has at least two \underline{\textbf{layers}} of meaning.}

$\bullet$  \textit{the mind functions on many \underline{\textbf{strata}} simultaneously.}
Given the above observations, we categorize the measurement of Eq.~\ref{equation:s1} into two situations, namely the same sense expressed in the same word ($\texttt{Sim}_\texttt{ss}$) and the same sense expressed in different words ($\texttt{Sim}_\texttt{ds}$), using the equations below:

\begin{equation}
\begin{split}
\texttt{Sim}_\texttt{ss}=\frac{\sum_{i,j}{\rm Cos}(t_i,t_j)}{{\rm Count}(t_i,t_j)},t_i=t_j,s_i=s_j \\
\texttt{Sim}_\texttt{ds}=\frac{\sum_{i,j}{\rm Cos}(t_i,t_j)}{{\rm Count}(t_i,t_j)},t_i\neq t_j,s_i=s_j 
\label{equation:s3}
\end{split}
\end{equation}

\begin{table}[t]
	\small
	\centering
	\begin{tabular}{c|c|c|c|c}
		\hline 
		{\bf{Model}}& \textbf{$\texttt{Sim}_\texttt{ss}$} & \textbf{$\texttt{Sim}_\texttt{ds}$}& \textbf{$\texttt{Sim}_\texttt{rand}$} & \textbf{$\Delta_\texttt{ss,rand}$}\\
		\hline
		ELMo &  0.45 & 0.29&0.17&	$\downarrow$    0.28\\
		
		\hline
		BERT &  0.67 & 0.62 &0.47&$\downarrow$    {0.20}\\
		\hline
		SenseBERT &  0.68 & 0.64 &0.35&$\downarrow$    \textbf{0.33}\\
		\hline
		RoBERTa &0.91 & 0.85&0.80&$\downarrow$    0.11\\	
		
		\hline
		DeBERTa &0.79 & 0.72&0.62&$\downarrow$	0.17\\	
		\hline
		
		XLNet & 0.93 & 0.92&0.92&$\downarrow$	0.01\\
		\hline
		GPT2 &  \textbf{0.99} & \textbf{0.98}&\textbf{0.97}&$\downarrow$	0.02\\
		\hline
	\end{tabular}
	\caption{Comparison between the sense-level similarity and random baseline.}
	\label{table:overall}
\end{table}

\begin{table}[t]
	\small
	\centering
	\begin{tabular}{c|c|c||c|c}
		\hline 
		{\bf{Model}}& \textbf{$\texttt{Sim}_\texttt{ss}$} &\textbf{$\texttt{Sim}^\texttt{mask}_\texttt{ss}$} & \textbf{$\texttt{Sim}_\texttt{ds}$}&  \textbf{$\texttt{Sim}^\texttt{mask}_\texttt{ds}$}\\
		\hline
		
		BERT &  0.67 &0.64($\downarrow$.03)& 0.62 &0.58($\downarrow$.04)\\
		\hline
		SenseBERT &  0.68  &0.65($\downarrow$.03)& 0.64&0.59($\downarrow$.05)\\
		\hline
		RoBERTa &0.91 &0.88($\downarrow$.03)& 0.85&0.82($\downarrow$.03)\\	
		
		\hline
		DeBERTa &0.79 &0.77($\downarrow$.02)& 0.72&0.69($\downarrow$.03)\\	
		
		\hline
	\end{tabular}
	\caption{Comparison of results between the masked and unmasked token representations.}
	\label{table:mask-results}
\end{table}

\textbf{Masked Words in the Same Sense.} For BERT and its variants (SenseBERT, RoBERTa, and DeBERTa), they use masked language modeling to predict the masked word during pre-training. Intuitively, the representation of token $\texttt{[MASK]}$ can also reflect the sense-level information by its context. Similar to Eq.~\ref{equation:s3}, we considered the comparison of such contextualized embeddings without the cues of the given token:
\begin{equation}
\begin{split}
\texttt{Sim}^\texttt{mask}_\texttt{ss}=\frac{\sum_{i,j}{\rm Cos}(\widetilde{t}_i,\widetilde{t}_j)}{{\rm Count}(t_i,t_j)},t_i=t_j,s_i=s_j \\
\texttt{Sim}^\texttt{mask}_\texttt{ds}=\frac{\sum_{i,j}{\rm Cos}(\widetilde{t}_i,\widetilde{t}_j)}{{\rm Count}(t_i,t_j)},t_i\neq t_j,s_i=s_j 
\label{equation:s4}
\end{split}
\end{equation}
where $\widetilde{t}_i$ and $\widetilde{t}_j$ are $\texttt{[MASK]}$ symbols that replace the origin ${t}_i$ and ${t}_j$.

\textbf{Random Baseline.} The above scores reflect the context sensitivity of the same word sense across models in absolute cosine similarities. However, for some CWE models, the general cosine similarity between arbitrary vectors can be large. In order to understand both the absolute distance and the relative difference, we also randomly sample $N$ = 10000 words and compute the average similarity among them as our baseline, and we define $\Delta_\texttt{ss,rand}$ as the difference between $\texttt{Sim}_\texttt{ss}$ and $\texttt{Sim}_\texttt{rand}$:
\begin{equation}
\begin{split}
\begin{aligned}
\texttt{Sim}_\texttt{rand} &= \frac{2}{N(N-1)}\sum_{i,j}{\rm Cos}(t_i,t_j) \\
\Delta_\texttt{ss,rand} &= \texttt{Sim}_\texttt{ss} - \texttt{Sim}_\texttt{rand}
\label{equation:rand}
\end{aligned}
\end{split}
\end{equation}

\subsection{Probing Method}
Cosine similarity has been widely used for measuring semantic similarity~\cite{miko2013a,sbert}, however, it only offers one global perspective of the embeddings. As a result,  we also consider adding a simple linear or MLP layer upon CWE for probing~\cite{conneau-etal-2018-cram,Nelson} to show the local or transformed features. In particular, we build a sense equivalent judging task by constructing 20,000 token pairs with sense annotations for training and 2,000 held-out samples for evaluation. We optimized the model by only finetuning the linear layer or MLP parameters while keeping the CWE fixed:
\begin{equation}
\begin{aligned}
P(y|t_i,t_j) &= {\rm softmax}(W_0([h_i;h_j])), {\rm or} \\
P(y|t_i,t_j) &= {\rm softmax}(W_1g(W_2([h_i;h_j]))) 
\end{aligned}
\end{equation}
where $W_0$, $W_1$ and $W_2$ are parameters, $g$ is the ReLU function and $[h_i;h_j]$ is the concatenation operation. $y$ is the binary distribution indicating the sense labels $s_i$ and $s_j$ are equal or not.  	

\section{Experimental Results}
\label{section:results}

\begin{figure*}[!t]
	
	\centering
	\includegraphics[scale=0.28]{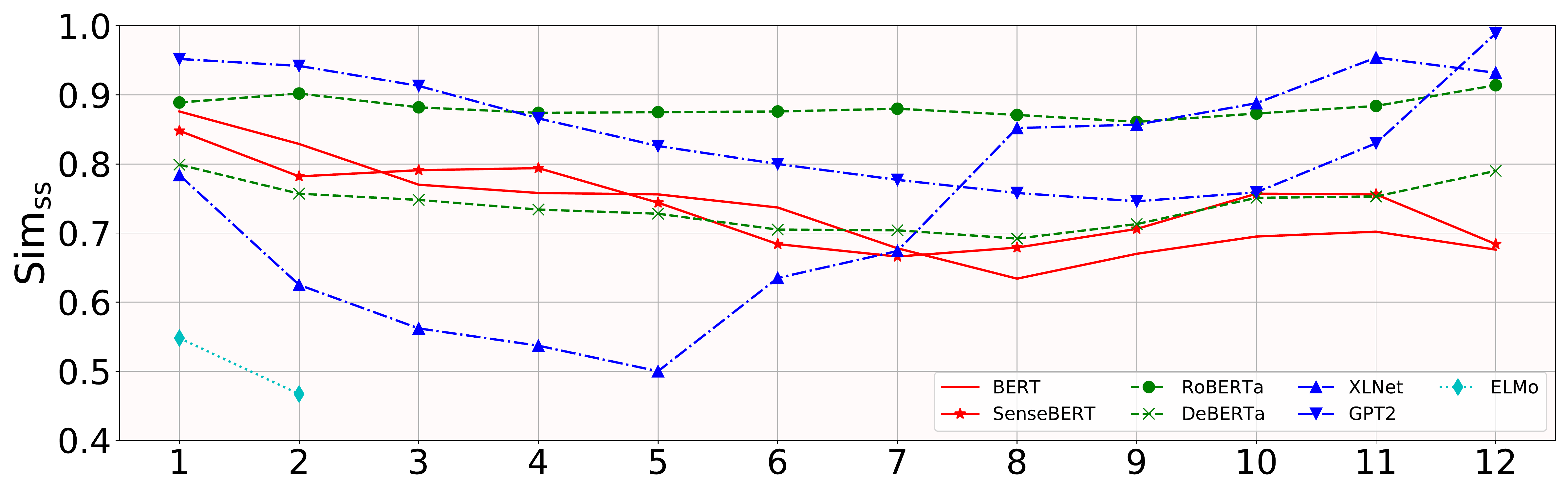}
	\caption{Layer-wise cosine similarity between the \textit{same} word in the \textit{same} sense. }
	\label{figure:sameword_sense}
	
\end{figure*}
\begin{figure*}[!t]
	\centering
	\includegraphics[scale=0.28]{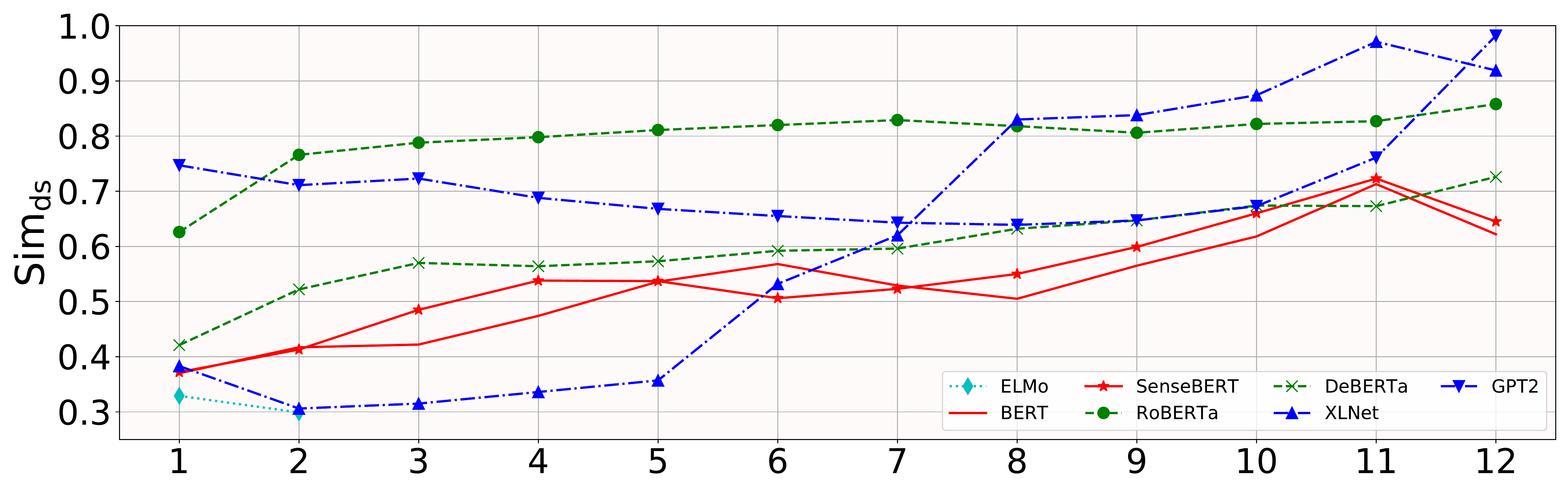}
	\caption{Layer-wise cosine similarity between the \textit{different} word in the \textit{same} sense.}
	\label{figure:_diffword_samesense}
\end{figure*}

\begin{table}[t]
	\centering
	\small
	\begin{tabular}{c|c|c}
		\hline 
		\textbf{Model} & \textbf{Linear} & \textbf{MLP} \\
		\hline 
		{ELMo} & 63.60 & 86.00\\
		\hline 
		{BERT} & 63.45 &85.95\\
		\hline 
		{SenseBERT} & \textbf{64.90}& \textbf{87.90}\\	
		\hline 
		{RoBERTa} & 63.70 &81.20\\
		\hline 
		{DeBERTa} & 63.75 & 86.25\\
		\hline 
		{XLNet} & 61.85 &60.05\\
		\hline 
		{GPT2} & 60.35 & 59.75\\
		\hline
	\end{tabular}
	\caption{Results of sense equivalent probing experiments, reported by accuracy.}
	\label{table:probe}
\end{table}

Table~\ref{table:overall} and Table~\ref{table:mask-results} show the results of cosine similarity using the CWEs from the last layer.  Overall, both $\texttt{Sim}_\texttt{ss}$ and $\texttt{Sim}_\texttt{ds}$ are relatively large compared with  $\texttt{Sim}_\texttt{rand}$, showing that CWE keeps a high level of sense-wise similarity across contexts, which can be a useful characteristic. The values of $\texttt{Sim}_\texttt{ss}$ are higher than those of $\texttt{Sim}_\texttt{ds}$, which reflects the fact the same word is more likely semantically equivalent compared with different words. Table~\ref{table:probe} shows the results of  probing task, showing that CWE can also capture semantic features through simple linear or MLP transformation.

\subsection{Comparison Between CWE Models}

\textbf{Different models show different similarities. }
In Table~\ref{table:overall}, the models show large variations in vector similarities. ELMo gives the lowest $\texttt{Sim}_\texttt{ss}$, $\texttt{Sim}_\texttt{ds}$, and $\texttt{Sim}_\texttt{rand}$ of 0.45, 0.29 and 0.17, respectively. In contrast, Transformer-based models give higher scores. The reasons can be that: 1) The dimension of embedding space (1024 vs. 768) is different; 2) Transformer-based models use attention mechanism rather than recurrent state transition for information exchange, making the token interaction direct and the output embeddings more similar.

\begin{table*}[t]
	\centering
	\small
	\scalebox{0.98}{
		\begin{tabular}{l}
			\hline 
			\hline 
			\centerline{\textbf{\textit{Maximum} Similarity Scores among BERT Representations }}\\
			\hline 
			\textbf{Microscopically: \textsc{bn00116251r}} (By using a microscope)  \hfill{\textbf{$\texttt{Sim}_\texttt{ss}$ = 0.996}} \\
			\hline 
			1. \underline{\textit{Microscopically}}, there was emphysema, fibrosis, and vascular congestion. \\
			2. \underline{\textit{Microscopically}}, there was hyperemia of the central veins, and there was some atrophy of adjacent parenchyma. \\
			3. \underline{\textit{Microscopically}}, the mucosa of the stomach showed extensive cytolysis and contained large numbers of ...\\
			\hline 
			\textbf{Petitioner: \textsc{bn00061827n}} (One praying humbly for something)  \hfill{\textbf{$\texttt{Sim}_\texttt{ss}$ = 0.991}} \\
			\hline 
			1. \underline{\textit{Petitioner}}, who claims to be a conscientious objector, was convicted of violating 12(a) of ... \\
			2. \underline{\textit{Petitioner}} was found guilty and sentenced to 15 months' imprisonment. \\
			3. \underline{\textit{Petitioner}} also claimed at trial the right to inspect the original Federal Bureau of Investigation reports to ... \\
			\hline 	
			\textbf{Second: \textsc{bn00116918r}} (In the second place)  \hfill{\textbf{$\texttt{Sim}_\texttt{ss}$ = 0.985}} \\
			\hline 
			1. \underline{\textit{Second}}, in a competitive market, the customer feels his weight and throws it around. \\
			2. \underline{\textit{Second}}, the upper portion permits comparison of maturity levels of an equal number of growth centers from ...\\ 
			3. \underline{\textit{Second}}, even if the characteristic polynomial factors completely over F into a product of polynomials of degree 1, ...\\
			\hline 	
			\hline

			\centerline{\textbf{\textit{Minimum} Similarity Scores among BERT Representations }}\\
			\hline 
			\textbf{pictorial: \textsc{bn00108586a}} (Pertaining to or consisting of pictures)  \hfill{\textbf{$\texttt{Sim}_\texttt{ss}$ = 0.384}} \\
			\hline 
			1. The affixed elements of collage were extruded, as it were, and cut off from the literal \underline{\textit{pictorial}} surface to form a bas-relief. \\
			2. It resembles, too, pictures such as Du\^rer and Bruegel did, in which all that looks at first to be solely \underline{\textit{pictorial}} proves on ...\\
			3. Flatness may now monopolize everything, but ... at least  an optical if not, properly speaking, a \underline{\textit{pictorial}} illusion. \\
			
			\hline 
			\textbf{bear: \textsc{bn00083221v}} (Have)  \hfill{\textbf{$\texttt{Sim}_\texttt{ss}$ = 0.385}} \\
			\hline

			1. In any given period piece the costumes, bric-a-brac, vehicles, and decor, \underline{\textit{bear}} the stamp of unimpeachable authenticity. \\
			2. The senses in each counterpart \underline{\textit{bear}} the impression only of phenomena that share its own frequency, whereas those ...\\
			3. Bob Carroll may not \underline{\textit{bear}} quite as close a physical resemblance to LaGuardia as Tom Bosley does, but I was amazed at ...\\

			\hline 	
			\textbf{neurotic: \textsc{bn00107256a}} (Characteristic of or affected by neurosis)  \hfill{\textbf{$\texttt{Sim}_\texttt{ss}$ = 0.386}} \\
			\hline 
			
			1. ... Wolpe thought that he could utilize the feeding pain antagonism to inhibit the \underline{\textit{neurotic}} symptoms through feeding.\\
			2. We pointed out that emotional excitement may lead to psychosomatic disorders and \underline{\textit{neurotic}} symptoms, particularly ...\\
			3. In view of the important role which emotional disturbances play in the genesis of \underline{\textit{neurotic}} and psychotic disorders ...\\
			\hline 
			
			\hline
	\end{tabular}}
	\caption{Words and senses in our dataset that give the maximum or minimum similarity using BERT representation.}
	\label{table:example_eg}
\end{table*}	

\textbf{CWE models encode word sense knowledge.} ELMo and BERT show a relative large difference between $\texttt{Sim}_\texttt{ss}$ and $\texttt{Sim}_\texttt{rand}$, reflecting the distinct spatial distribution according to different word senses~\cite{vis}. Our findings align with Garí Soler and Apidianaki~\cite{mono-poly}, who find a similar gap in BERT representation and show BERT representation can reflect the polysemous level. SenseBERT is trained with external word sense prediction and gives the largest difference, showing the usefulness of sense-level supervision signal during pre-training.

\textbf{Embedding space of auto-regressive models are highly anisotropic.} For XLNet and GPT2, the difference between $\texttt{Sim}_\texttt{ss}$ and $\texttt{Sim}_\texttt{rand}$ ($\Delta_\texttt{ss,rand}$) are only 0.01 and 0.02, respectively. $\texttt{Sim}_\texttt{ds}$ is even slightly lower than $\texttt{Sim}_\texttt{rand}$ in XLNet. These can result from the highly anisotropic distribution of word vectors in the final layers, as has been observed by Ethayarajh~\cite{ethayarajh-2019-contextual}, especially in GPT2. In this case, the word vectors assemble in a narrow cone rather than being uniform in all directions. We find a similar phenomenon in XLNet representations and hypothesize that it is due to the auto-regressive language modeling objective.

\textbf{Both token and context are non-negligible for CWE.} Table~\ref{table:mask-results} shows the results of the masked token. Compared with the unmasked representation, the similarities drop consistently, showing the importance of the token feature for the final CWE. However, the difference is not significant, ranging from -0.05 to -0.02. The results of $\texttt{Sim}^\texttt{mask}_\texttt{ds}$ are still larger than the random baseline, showing that the sentential information is also utilized in models.

\textbf{Probing results are consistent with cosine similarity.} Table~\ref{table:probe} shows the results of the probing task. We find that the performance of using a linear classifier is around 60\% accuracy. The results are improved much when using MLP for ELMo, BERT, and its variants, reaching 80\% or more. As for XLNet and GPT2, the results are both low, where CWE distribution is extremely non-uniform. Overall, the performance is highly related to the cosine similarity based $\Delta_\texttt{ss,rand}$ in Table~\ref{table:overall}, showing that the similarity gap can reflect the capability of encoding sense level features. For brevity, we use the cosine similarity metric for further analysis in the remaining of this paper.

\subsection{Influence of Model Layers}

Besides the final layer that is considered to encode more semantic-related features, we draw the layer-wise results in Figure~\ref{figure:sameword_sense} and Figure~\ref{figure:_diffword_samesense}, respectively. 

\textbf{Features are encoded hierarchically in models.} In both figures, sense-wise similarity varies according to different layers. As seen in Figure~\ref{figure:sameword_sense}, $\texttt{Sim}_\texttt{ss}$ in the bottom layers are relatively large compared with that in the middle layer, while the values are increasing in the higher layers (except for a slight drop in the final layers of BERT, SenseBERT, and XLNet). These are related with recent findings about knowledge learned in CWE. For example, ~\citet{belinkov-etal-2020-linguistic} find that morphology is learned at the lower layer of CWE, the large $\texttt{Sim}_\texttt{ss}$ come from the same word spellings. For the middle layers, BERT learns more non-local linguistic knowledge such as syntax~\cite{Goldberg2019AssessingBS}, making the sense-wise similarity lower. For higher layers, the word representation is highly contextualized~\cite{ethayarajh-2019-contextual} and encodes more semantic knowledge~\cite{jawahar}, thus the sense-level similarity increase.

\textbf{Deep layers are required to extract contextual meaning.} As shown in Figure~\ref{figure:_diffword_samesense}, for the different words in the same sense, almost all the models have relatively low similarity scores (0.3$\sim$0.7) in the bottom layers, where the embeddings are not highly contextualized.  The $\texttt{Sim}_\texttt{ds}$ values generally increase as the layer moves higher, and reaches high values (0.6$\sim$0.9) in the top layer representations, demonstrating that CWE encodes context information and recognizes the shared word sense. 

\textbf{Different models show different trends}, which may be because of the different training settings such as training corpus and  objective. For example, in Figure~\ref{figure:sameword_sense}, RoBERTa is more stable while XLNet shows the most variance according to model layers, ranging from 0.5$\sim$0.9. The reason can be the specific pre-training objective in XLNet, where the context is permuted during training.

\subsection{Case Study}

We select BERT and show the word/senses (appearing at least five times) that contribute the most and least to $\texttt{Sim}_\texttt{ss}$ in Table~\ref{table:example_eg}. \textbf{First}, three words \textit{``Microscopically''}, \textit{``Petitioner''}, and \textit{``Second''} give the highest similarity. They have a relatively stable meaning. In contrast, the words ``\textit{pictorial}'', ``\textit{bear}'', and ``\textit{neurotic}'' are the lowest-ranked, which are more polysemous. As in the dataset we used, the number of senses of words \textit{``Microscopically''}, \textit{``Petitioner''}, \textit{``Second''}, \textit{``pictorial''}, \textit{``bear''}, \textit{``neurotic''} are 1, 1, 2, 2, 12 and 3, respectively. \textbf{Second}, contexts for the latter three words are generally longer and more complex, such contextualization may have an effect on the word representations.  \textbf{Third}, interestingly, the  three words with the largest $\texttt{Sim}_\texttt{ss}$ appear in the sentence start position, yet the three  with the lowest $\texttt{Sim}_\texttt{ss}$ appear in different and irregular positions of contexts.

The above case suggests that both intrinsic and extrinsic characteristics have an influence on the cross-context stability of word embeddings. We analyze the influence of different types of words (\S\ref{sec:51}) and contexts (\S\ref{sec:52}) accordingly. We also investigate if there is a position bias for high similarity for the first word, or it just happens by chance (\S\ref{sec:53}).

\begin{figure*}[!t]
	\centering
	\includegraphics[scale=0.32]{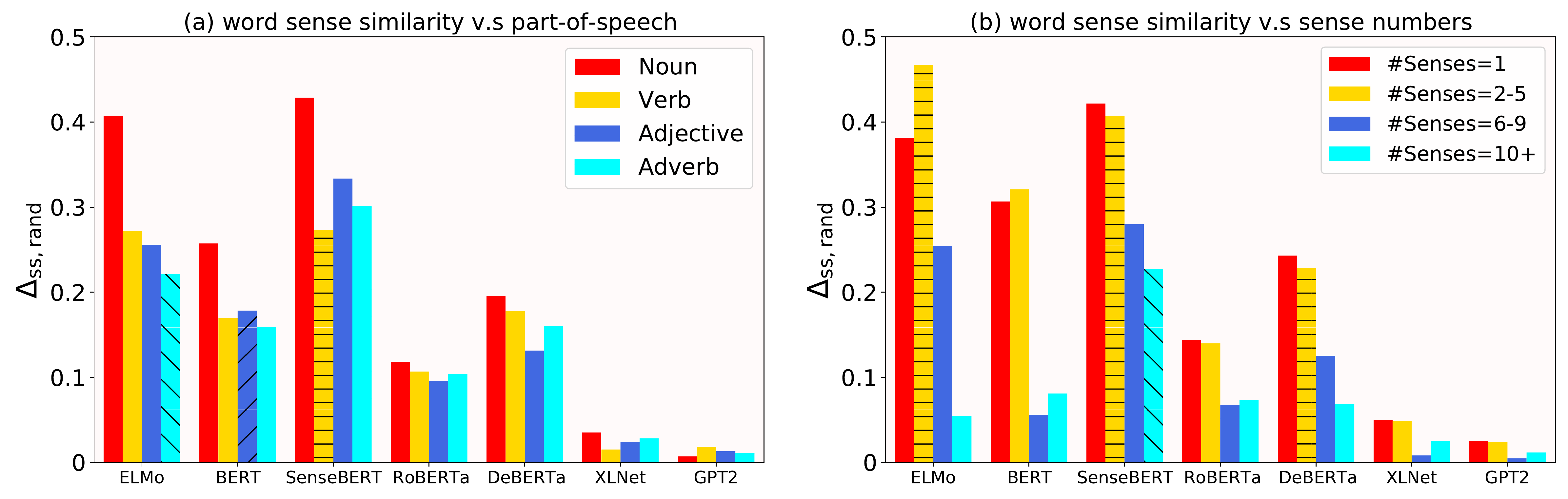}
	\caption{Influence of word to the similarity w.r.t (a) part-of-speech, and (b) number of senses.}
	\label{figure:word_type}
\end{figure*}

\begin{figure*}[!t]
	\centering
	\includegraphics[scale=0.32]{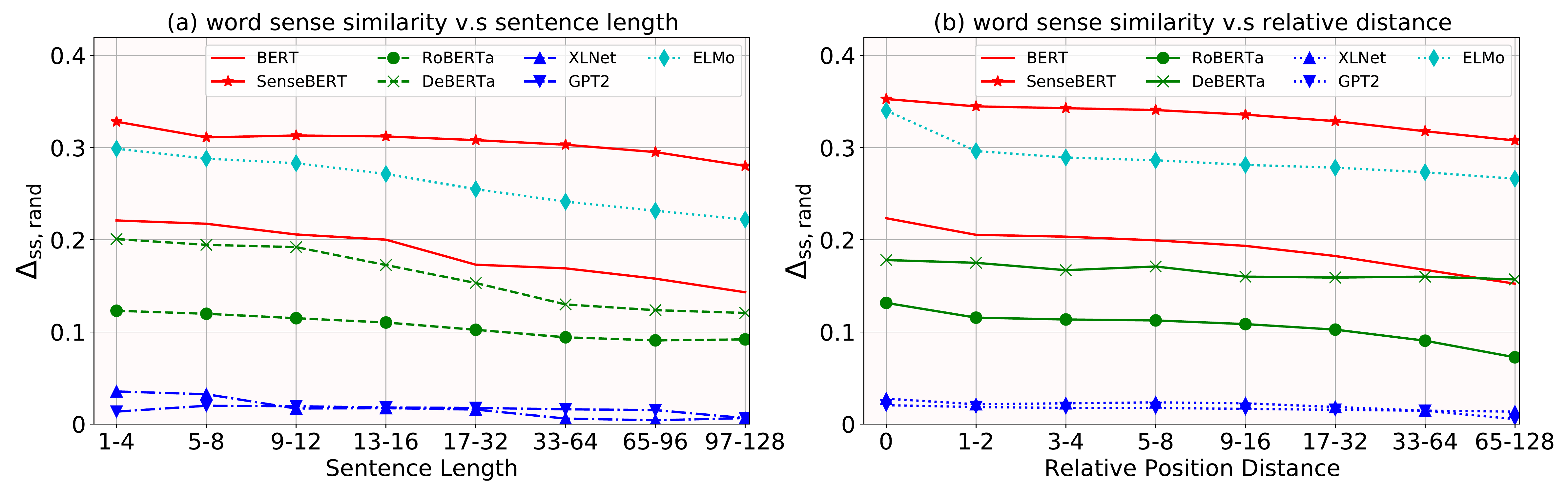}
	\caption{Influence of word to the similarity w.r.t sentence length, and (b) relative position distance.}
	\label{figure:context_type}
\end{figure*}

\section{Analysis and Discussion}
For word attribution, we investigate the influence of part-of-speech and the number of sense, given that (1) word senses are related to the PoS by definition; (2) the number of senses decide the  polysemous degree of a word. With regard to contextual-level influences, we are interested in the sentence length and the word position in sentence, for both a semantic variation is undesired.

\subsection{Influence of Word Type}
\label{sec:51}

\noindent\textbf{The Part-of-Speech.} Similar to Eq.~\ref{equation:s3}, we compute the cosine similarity according to different PoS categories:
\begin{equation}
\texttt{Sim}_{\texttt{ss}}[\texttt{PoS}] =\frac{\sum_{i,j}\texttt{Cos}(t_i,t_j)}{\mbox{Count}(t_i,t_j)}
\label{eq:eqpos}
\end{equation}
where $s_i$ and $s_j$ are both belong to one PoS among of the noun, the verb, the adjective or the adverb.

The results are shown in Figure~\ref{figure:word_type}(a). For GPT2 and XLNet, the values are small and the difference between PoS categories is small.  The differences for the other five models are larger according to different PoS. In general, the similarity of nouns is the largest across all models, which shows that the CWE of nouns are more consistent.  

\noindent\textbf{The Number of Senses.} We also compute the cosine similarity according to different numbers of word senses:

\begin{equation}
\texttt{Sim}_{\texttt{ss}}[\texttt{\#Senses}] =\frac{\sum_{i,j}\texttt{Cos}(t_i,t_j)}{\mbox{Count}(t_i,t_j)}
\label{eq:eqsenses}
\end{equation}
where the sense number of token $t$ (\texttt{\#Senses}) varies from 1 to 10 or more.

Figure~\ref{figure:word_type}(b) shows the $\Delta_\texttt{{ss,rand}}$ according to different numbers of senses. For all models, the values are higher for words with fewer senses (1 and 2$\sim$5) and decrease sharply as the number of senses increases (6$\sim$9 and 10+). The similarity value is also related to the granularity of defined word senses. For those common words with 10+ senses, such as the word ``\textit{take}'' labeled with 58 senses in the dataset, some of the senses can be easily confused, leading to the difficulty for sense understanding via CWE.

Each sense belongs to one PoS by definition, and we calculate the average number of senses of a word against each PoS. The results for nouns, verbs, adjectives, and adverbs are 1.35, 1.83, 1.46, and 1.24, respectively. The number of senses is relatively small for nouns, which can explain its high consistency. Adverb has got the lowest number of senses but also has the lowest word frequency in corpora. This makes adverbs less pre-trained generally, thus the results are similar to verbs and adjectives.

\subsection{Influence of Context}
\label{sec:52}

\noindent\textbf{The Sentence Length.} 
We compute the the similarity according to different sentence length and position distances:

\begin{equation}
\texttt{Sim}_{\texttt{ss}}[\texttt{SentLen}] =\frac{\sum_{i,j}\texttt{Cos}(t_i,t_j)}{\mbox{Count}(t_i,t_j)}
\label{eq:SentLen}
\end{equation}
where $t_i\in t_1..t_i..t_{N_i}$ and $t_j\in t_1..t_j..t_{N_j}$. $N_i, N_j$ are in the same length range.

Figure~\ref{figure:context_type}(a) shows the values of $\Delta_\texttt{{ss,rand}}$ regarding different sentence lengths. The largest similarity arises between words appearing in relatively short sentences. As the sentence becomes longer, the similarity decreases and  gradually stabilized. This shows that the CWE of words is influenced by the sentential content, the sense-level knowledge is easier recognized and more distinguishable in short sentences. However, the CWE becomes more changeable when being more contextualized, thus the sense-wise consistency is prone to getting lost.

\noindent\textbf{The Relative Distance.} 

\begin{equation}
\texttt{Sim}_{\texttt{ss}}[\texttt{Dist}] =\frac{\sum_{i,j}\texttt{Cos}(t_i,t_j)}{\mbox{Count}(t_i,t_j)}
\label{eq:PosDist}
\end{equation}
where the relative distance between $t_i$ and $t_j$ (\textit{i.e.}, $|i-j|$) is in the same distance range.

The results for word pairs in different position distances are shown in Figure~\ref{figure:context_type}(b). Overall, for ELMo, word pairs in the same position  (\textit{i.e.}, the relative distance is zero) give larger similarities. As the distance becomes longer, the similarity decreases. This may be because words with similar positions tend to share more similar context, thus the semantic relationship tends to be close.  In contrast, a larger distance means different contextualized information, making the variation of CWE. BERT, SenseBERT, and RoBERTa use \textit{absolute} position embeddings, where word embeddings in the same positions are concatenated with the same position embeddings as inputs, making the representations more similar. However, DeBERTa uses \textit{relative} position embeddings during the attention interaction, showing 
little similarity variance against relative distance.

\subsection{Position Bias for the First Word}
\label{sec:53}
According to the findings in Table~\ref{table:example_eg} and Figure~\ref{figure:context_type}(b), we find that there exists a position bias for the first word representation, where the similarity is irregularly higher than other positions.

\begin{figure*}[t!] 
	\centering 
	\includegraphics[scale=0.3]{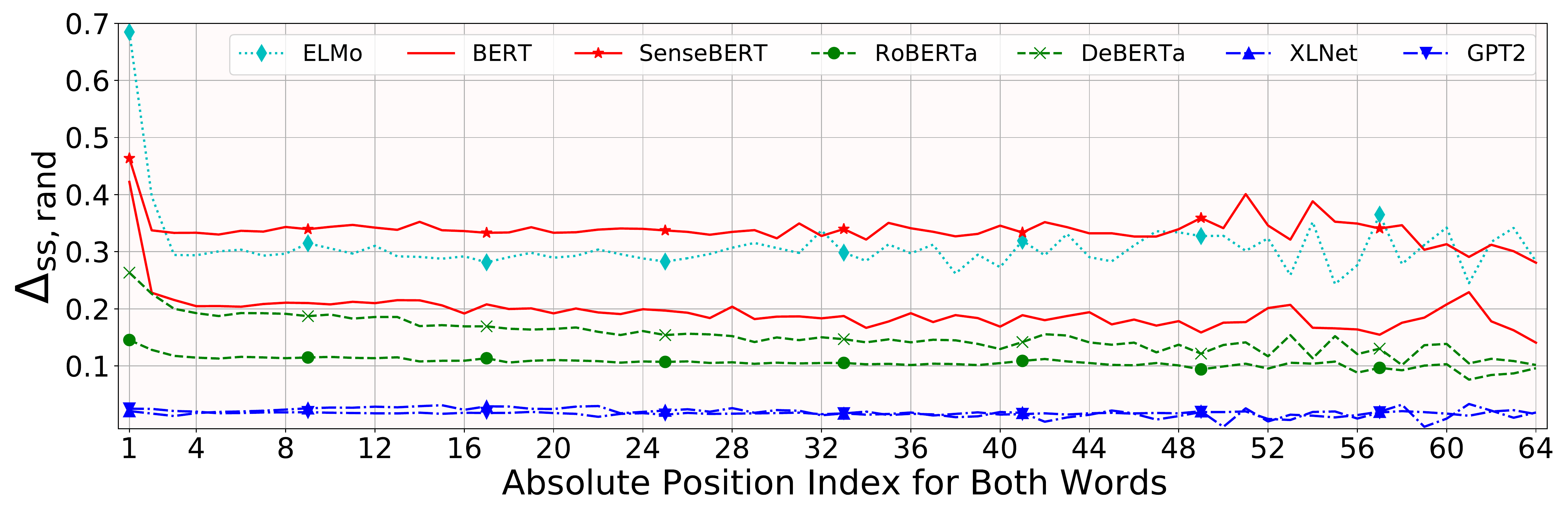}
	\caption{Similarity scores between words in the same positions vs. different position indices. }
	\label{figure:vis_positionsame}
\end{figure*}

\noindent\textbf{Part-of-Speech vs. Position.} We first check the distribution relationship between part-of-speech and position. Although the similarity for nouns is larger verbs, as shown in Table~\ref{table:composition}, the total number are similar (36.0\% vs. 36.7\%) and there is no obvious correlation between part-of-speech and positions, the distribution for the first words are normal as other words. This show that the position bias is not due to the imbalance distribution of part-of-speech in each position.

\noindent\textbf{Similarity vs. Position.} We set relative position distance as zero and measure the sense similarity at different position indices. The results are shown in Figure~\ref{figure:vis_positionsame}. Intuitively, the values of word similarity should be position-insensitive. However, we find that, except for XLNet and GPT2, the first few positions give the largest similarity. The values go down as the position moves forward, reaching a relatively stable value soon. 

\begin{table}[t]
	\small
	\centering
	\begin{tabular}{c|c|c|c|c|c}
		\hline 
		\multirow{2}{*}{\textbf{Part-of-Speech}}& \multicolumn{5}{c}{\textbf{Distribution(\%) at Position Index}}\\
		\cline{2-6}
		& 1 & 2& 3 & 5 & 10\\
		\hline
		Nouns (36.0\%) &  3.0 & 6.1&4.2&4.9 & 3.5\\
		
		Verbs (36.7\%) &  4.2 & 8.6& 8.2 & 5.1&  3.1  \\
		
		Adjective (18.8\%) &  13.8 & 16.8&5.1&   4.4&  2.5\\
		
		Adverb (8.5\%) & 9.3 & 3.8&8.7&  6.3& 2.9\\
		
		\hline
	\end{tabular}
	\caption{Word distribution of different part-of-speech and some corresponding position indices.}
	\label{table:composition}
\end{table}

The reason can include: \textbf{First}, the CWE of the sentence beginning word from ELMo gives the highest similarity because they share the same information before the first token (conveyed by hidden states $\overrightarrow{h_{\langle s \rangle}}$ of the sentence beginning symbol \textit{$\langle s \rangle $}), which serves half hidden states (together with the context information in the backward direction from the second word $\overleftarrow{h_2}$) for the generation of the final vector. \textbf{Second}, for BERT-related models, the first token shows higher similarity, which may be because the attention mechanism inside Transformers can encode certain patterns. For example, the words tend to focus on the [CLS] token directly or the previous tokens in certain attention heads~\cite{clark-etal-2019-bert}. Further, words can have strong local dependencies~\cite{pmlr-v97-gong19a}, which makes the CWE to be position biased, especially for the sentence beginning words that share the [CLS] token nearby.

\noindent\textbf{Verification by Prompting.} To further explore the influence of position, we verify it by adding three designed prompts below for each sentence [X]. In this way, the word attribution (e.g, sense and part-of-speech) in the original sentence keep steady, but the position for each word will shift, and the local context for the sentence beginning word is enriched, relying less on the [CLS] token. These prompts are:

\begin{table}[t]
	\small
	\centering
	\begin{tabular}{c|c|c|c|c|c}
		\hline 
		\multirow{2}{*}{\textbf{Data}}& \multicolumn{5}{c}{\textbf{$\Delta_\texttt{ss,rand}$ at each position index }}\\
		\cline{2-6}
		& 1 & 2-4& 5-8 & 9-16 & 17+\\
		\hline
		Original &  0.422 & 0.221&0.206&   0.209&   0.196  \\
		
		\hline
		Prompt 1 &  0.320 & 0.243&0.216&   0.216&   0.205\\
		%\cdashline{1-6}
		Prompt 2 &  0.311 & 0.258&0.230&   0.228&   0.214\\
		%\cdashline{1-6}
		Prompt 3 &  0.314 & 0.259&0.226&   0.224&   0.210\\
		\hline
		{\textbf{$\Delta$ (Avg.)}} & $\downarrow$\textbf{0.11}&   $\uparrow$0.03&$\uparrow$0.02&   $\uparrow$0.01&   $\uparrow$0.01\\
		\hline
	\end{tabular}
	\caption{Variations of similarity between words in the same positions by adding different prompts.}
	\label{table:prompt}
\end{table}

\begin{figure*}[t!] 
	\centering 
	\includegraphics[scale=1.6]{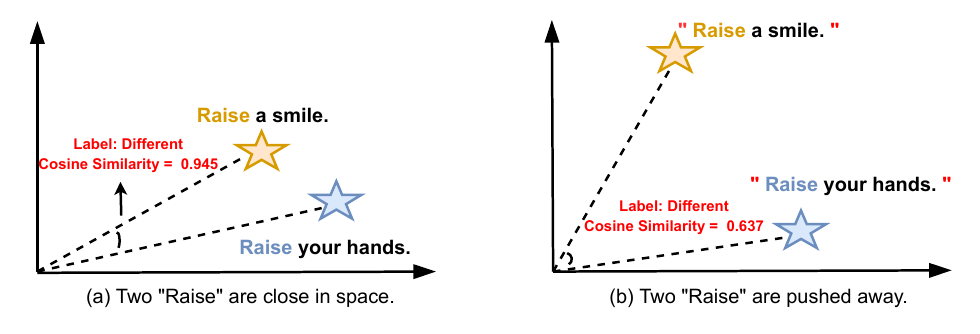}
	\caption{An illusion of position bias in CWE. (a) two words ``Raise" at the first position are extremely close, but they have different meanings. (b) by adding a simple prompt (two quotation marks at the two ends of text), two embeddings are pushed away with larger distance. }
	\label{figure:unsup_wsd}
\end{figure*}

\begin{table*}[t]
	\small
	\centering
	\begin{tabular}{c|c|c|c|c|c|c|c|c|c|c|c|c}
		\hline 
		\multirow{2}{*}{\textbf{Model}}& \multicolumn{12}{c}{\textbf{Layer}}\\
		\cline{2-13}
		& 1 & 2& 3& 4 & 5& 6 & 7& 8& 9 & 10& 11 & 12 \\
		\hline
		{w/ original text}&\textbf{63.5}&\textbf{64.5}&\textbf{65.3}&65.7&65.9&65.8&65.5&65.4&65.6&65.3&64.5&66.9\\
		\hline
		{w/ prefix=``, suffix="}&63.0&63.9&64.7&65.2&65.4&65.3&65.1&65.2&65.6&65.7&64.9&67.7 (+0.8)\\
		\hline
		{w/ prefix=She said :}&63.2&64.2&65.1&\textbf{65.7}&\textbf{66.0}&\textbf{66.0}&\textbf{65.8}&\textbf{65.7}&\textbf{66.1}&\textbf{66.2}&\textbf{65.4}&68.1 (+1.2)\\
		\hline
		{w/ prefix=Document :}&62.9&63.8&64.7&65.1&65.2&65.1&64.8&64.8&65.0&65.2&64.8&\textbf{68.3 (+1.4)}\\
		\hline
	\end{tabular}
	\caption{ Results of using BERT for distance-based WSD in WiC dataset.}
	\label{table:unsup_wsd}
\end{table*}	

\noindent$\bullet$   \textbf{{Prompt1. }}\underline{\textit{``}} [X]. \underline{\textit{''}}

\noindent$\bullet$   \textbf{{Prompt2. }}\underline{\textit{She said :}} [X].

\noindent$\bullet$   \textbf{{Prompt2. }}\underline{\textit{Document :}} [X].

Table~\ref{table:prompt} shows the results. By adding different prompts, the similarity scores drop for the first word in the original sentence. For the other positions, the scores keep relatively stable with variations ranging from 0.01$\sim$0.03. This shows that the representations of the sentence beginning word are biased and context-sensitive, leading to the higher scores in Figure~\ref{figure:vis_positionsame} compared with other positions. We also illustrate an example of such interesting phenomenon in Figure~\ref{figure:unsup_wsd}.

\section{Position Bias Mitigation for Distance-based WSD}
In the above discussion, we show that the word representation of CWE is position biased, where the CWE of the first words in different contexts tends to be much closer than in other positions. In this section, we show that such bias may influence the performance of a certain downstream task and propose a simple way to mitigate it.

\subsection{Task definition and Datasets}
In particular, we choose the 2-way word sense disambiguation task using the Word-in-Context (WiC) dataset~\cite{pilehvar-camacho-collados-2019-wic}, which is also a sub-task included in the SuperGLUE benchmark~\cite{superglue}. Given two sentences and each sentence contain a target word, the task is to determine whether these two target words has the same contextualized meaning.

\subsection{Methods}
Since pre-trained models such as BERT has shown the capability for encoding semantics, we proposed a distance-based method by using BERT as our baseline. Formally, given two target words $t^1_i$, $t^2_j$ and two contexts $s^1=t^1_1, ..., t^1_i,...,t^1_M$ ,  $s^2=t^2_1, ..., t^2_j,...,t^2_N$, we assign the label to 1 (has same meaning) or 0 (has different meaning) according to their cosine distance between two CWEs:

\begin{equation}
\texttt{Label}(t^1_i, t^2_j) = \mathds{1} \       \llbracket{\rm Cos}(t^1_i, t^2_j) > {\rm T}\rrbracket 
\label{equation:unsup_wsd}
\end{equation}
where $\mathds{1}\llbracket\cdot\rrbracket $ is the indicator function, ${\rm T}$ is the best pre-specified threshold according to datasets.

The position bias show that the ${\rm Cos}(t^1_i, t^2_j)$ could be higher when $i=j=1$ for all word types $t$, thus the threshold  ${\rm T}$ could be higher than words in other positions. According to the experiments in previous section, we can add some prompt (prefix or suffix) to the original text without losing too much semantics, and make the positions for each tokens shift for a distance. We illustrate our method in Figure~\ref{figure:unsup_wsd}.

\subsection{Results}
The overall results are shown in Table~\ref{table:unsup_wsd}. Across all models, the accuracy increases from lower layers to higher layers, showing that again that higher layers encode more semantic-related information. The method using original text gives 66.9 accuracy in the last layer. Adding some prefix (and suffix) can improve the overall accuracy by 0.8$\sim$1.4.

\begin{table}[t]
	\small
	\centering
	\begin{tabular}{c|c|c}
		\hline 
		\multirow{2}{*}{\textbf{Model}}& \multicolumn{2}{c}{\textbf{Positions}}\\
		\cline{2-3}
		& $i=j=1$ & others\\
		\hline
		{w/ original text}&53.5&67.9\\
		\hline
		{w/ prefix=``, suffix="}&\textbf{61.9 (+8.4)}&68.1 (+0.2)\\
		\hline
		{w/ prefix=She said :}&59.2 (+5.7)&68.7 (+0.8)\\
		\hline
		{w/ prefix=Document :}&58.6 (+5.1)&\textbf{68.9 (+1.0)}\\
		\hline
	\end{tabular}
	\caption{ Results of using BERT (last layer) for distance-based WSD between $t^1_i$ and $t^2_j$.}
	\label{table:unsup_wsd_11}
\end{table}	

We also check the fine-grained results for different positions. As Table~\ref{table:unsup_wsd_11} shows, the accuracy for the first words are quite low (with only 53.5 by using the original text), showing that the CWEs for the first words have a low degree of distinction. As a comparison, the accuracy for the other positions is relatively high (about 67.9). By adding prefix, the accuracy for the original first words gets a large improvement (+5.1$\sim$+8.4), which shows the effectiveness of our method.

\section{Conclusion}
We investigated the sense-wise representations on contextualized word embeddings from seven pre-trained models. Results show that models can capture sense consistency well. However, influence from both intrinsic and extrinsic factors such as position bias also exists. Our work reveal some characteristics  as well as limitations of pre-trained language models and we hope it can help people better understand or improve language models from a semantic representation perspective.
\bibliography{anthology,custom}
\end{document}